\documentclass[letterpaper, 10 pt, conference]{ieeeconf}

\IEEEoverridecommandlockouts                              % This command is only needed if 
\usepackage{graphicx}
\usepackage{stmaryrd} % for pdf, bitmapped graphics files
\usepackage{epsfig} % for postscript graphics files
\usepackage{times} % assumes new font selection scheme installed
\usepackage{amsmath} % assumes amsmath package installed
\usepackage{amssymb}  % assumes amsmath package installed
\usepackage{float}
\usepackage{hyperref}
\usepackage{subcaption}
\usepackage{physics}
\usepackage{tikz}
\usepackage{pgfplots}
\usepackage{balance}
\usepackage{tabu}
\usepackage{multirow,booktabs}
\title{\LARGE \bf
Fast Loop Closure Detection via Binary Content
}
\author{Han Wang, Juncheng Li, Maopeng Ran and Lihua Xie% <-this % stops a space
\thanks{
The work is partially supported by Delta - NTU Corporate Lab under the NRF Corporate Lab @ University Scheme. 
}
\thanks{The authors are with the School of Electrical and Electronic Engineering,
Nanyang Technological University, 50 Nanyang Avenue, Singapore 639798.
        {\tt\small \{wang.han,elhxie,mpran\}@ntu.edu.sg, juncheng001@e.ntu.edu.sg}}%
}

\begin{document}
\maketitle
\thispagestyle{empty}
\pagestyle{empty}

%注意!!!!!!! At the beginning, we have to know what we want to do, and what is the advantage of our method what to compare, first of all, what do you want to present in the experiment result, compared to existing method, our advantage is we add a fast indexing to precise SURF deature indexing, so that the recall rate can be higher but at same speed with DBOW.
%什么是我们的方法的优点: 传统的bow只考虑到了特征点,但是这些特征点是无序的, 所以损失了重要的信息, 我们认为形状结构也是一个图像的重要组成部分, 并且图像结构所信息比特征点更简洁, 
%改了改了, 这篇文章我们就只说利用map加速传统特征点方法, 不要说无序的概念,留给iros

\begin{abstract}
Loop closure detection plays an important role in reducing localization drift in Simultaneous Localization And Mapping (SLAM). It aims to find repetitive scenes from historical data to reset localization. To tackle the loop closure problem, existing methods often leverage on the matching of visual features, which achieve good accuracy but require high computational resources. However, feature point based methods ignore the patterns of image, i.e., the shape of the objects as well as the distribution of objects in an image. It is believed that this information is usually unique for a scene and can be utilized to improve the performance of traditional loop closure detection methods. In this paper we leverage and compress the information into a binary image to accelerate an existing fast loop closure detection method via binary content. The proposed method can greatly reduce the computational cost without sacrificing recall rate. It consists of three parts: binary content construction, fast image retrieval and precise loop closure detection. No offline training is required. Our method is compared with the state-of-the-art loop closure detection methods and the results show that it outperforms the traditional methods at both recall rate and speed.
\end{abstract}

\section{INTRODUCTION}
Over the past decades, loop closure detection has become an important part of visual SLAM. In the early stage of development, SLAM only targeted at visual odometry which accumulates inevitable drifts. The result of navigation and mapping often fails in the long run. Later, it is found that graph based optimization can greatly correct drifting error with the help of loop closure detection and it becomes an essential component of modern SLAM \cite{williams2009comparison}. Nowadays, a full visual SLAM system consists of front-end and back-end. In the front-end, vision based SLAM runs visual odometry to estimate the frame-to-frame transition directly. However, visual odometry often has the problem of cumulative drift in real applications. In the back-end, loop closure partially resets localization to minimize transitional measurement error by matching current frame with historical data \cite{lowry2016visual}. Visual SLAM has been widely applied into many robotics fields such as cleaning robot, drone as well as autonomous cars. It has become a promising technique in robotics.
\begin{figure}[t]
\begin{center}
    \includegraphics[width=0.90\linewidth]{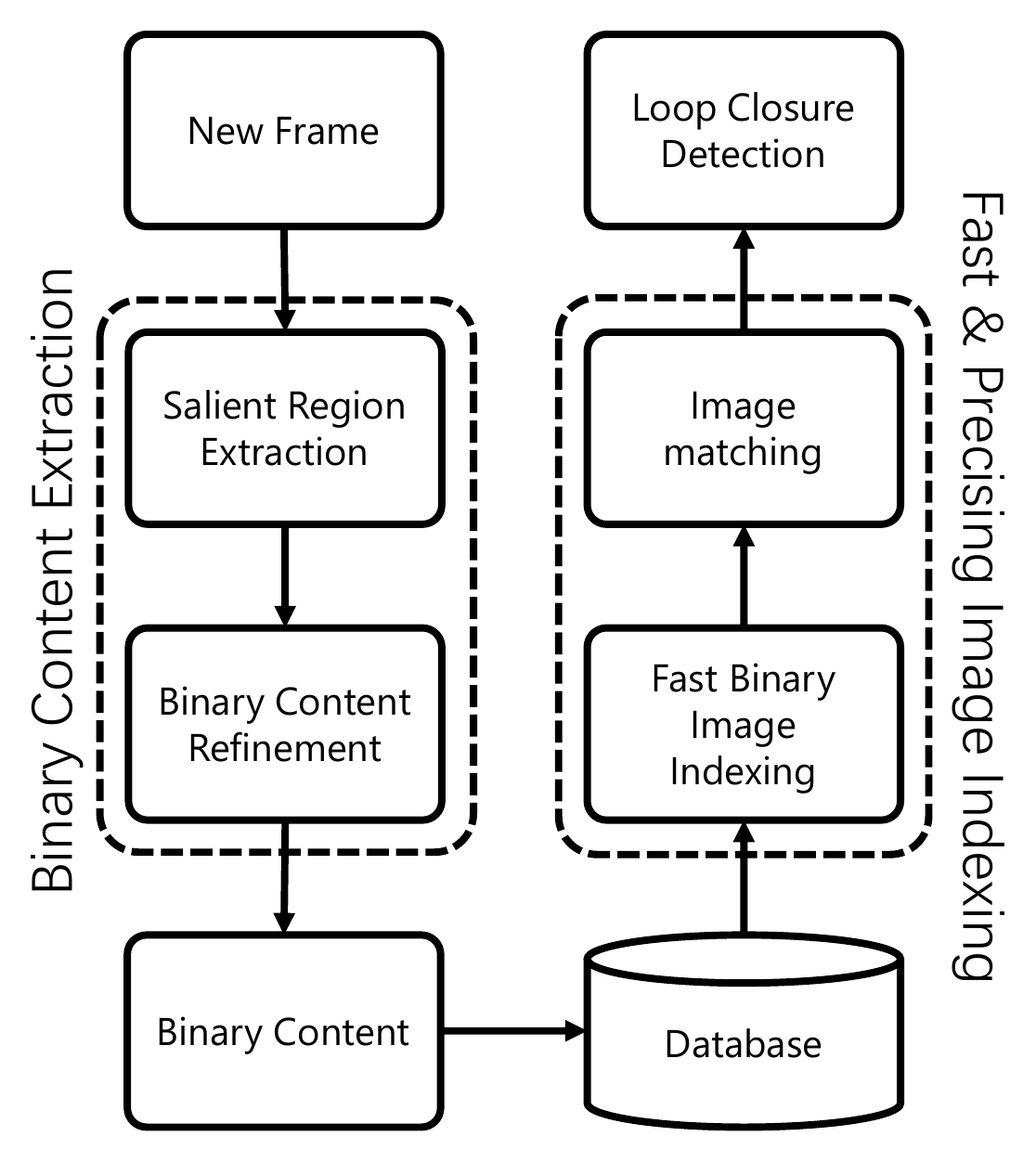}
\end{center}
\captionsetup{justification=centering}
\caption{Framework of Binary Content Based Loop Closure Detection.}
\label{fig: flowchart}
\end{figure}

The idea of loop closure detection is to find repetitive scenes from the historical data so that we can link two places together. The link between two places acts as an additional constraint to the mapping. After applying graph optimization, we can minimize the drifting error based on those constraints. Experiments have shown that loop closure detection can greatly improve the performance of SLAM \cite{williams2009comparison}. However, tackling this problem consists of many challenges; First of all, the database grows with time, meaning that the database size can be tremendous without proper compression. Secondly, the complexity of indexing grows proportionally with database size. Hence the requirement of computational resources also gradually increases. Lastly, two frames of same place taken at different timestamp may be slightly different due to variation of light condition, dynamic objects, etc. Therefore, loop closure detection still remains a challenging topic in visual SLAM.

Existing works on loop closure detection share the common idea of using hand-crafted feature points and feature descriptors, such as FAB-MAP\cite{cummins2008fab}, Bag of Visual Words (BoVW) \cite{galvez2012bags}, VLAD \cite{huang2016vlad} and Fisher Vector \cite{uchida2016image}. These methods extract feature points from image frame and translate them into descriptors. Then the descriptors are stored into database in sequence and we can simply tell where loop closure happens by comparing current descriptor and database. The number of comparison grows with time, hence in general the comparison of descriptors must be fast. However, there is always a trade-off between speed and precision. To achieve higher accuracy, it takes lots of computational resources. For example, in FAB-MAP \cite{cummins2008fab}, it takes 400 ms to extract SIFT features for a frame of size $640\times 480$ pixels on a normal computer. Image descriptors such as Fisher Vector contain high-order statistics so that it takes more time to process. In conclusion, existing methods leverage on creating accurate image descriptors but lack of satisfactory efficiency.

In this paper, we argue that extracting and comparing feature descriptors takes too much computational resources and becomes a burden to the processing system over a long run. Existing feature point based methods can achieve satisfactory recall rate results but are difficult to run in real time. In the meantime we note that the distribution of the objects or salient patterns is also an important information except from feature points. For a scene, the geometrical distribution of the objects as well as the shape of each object are usually unique and hence this can be used for loop closure detection. However, this information is not utilized in feature point methods. A good fact about the pattern information is that it does not involve any color information. And if we can express it in binary format, the speed can be improved. Hence we introduce this feature for loop closure detection, where the object distribution information is expressed as the binary content of image. Thereafter, we can verify the loop closure places by checking the similarity of the binary contents of two images. At the same time, we keep existing feature point method on top of binary content indexing to achieve high recall rate. The new framework consists of three parts: binary content construction, fast image retrieval and precise loop closure detection. It firstly introduces a binary map into loop closure detection to reduce the computational cost for indexing while applies precise image matching to guarantee precision. Compared to the existing methods, no offline training is required for our method. It is also proven that our method outperforms existing methods at both recall rate and speed. The main contributions of this paper are as follows:

\begin{itemize}
\item We propose a binary content based fast loop closure detection, which combines the advantage of both fast binary operation and traditional loop closure detection approach. 
\item The performance is greatly improved. Our method is much faster than existing methods without reducing recall rate and recall precision. 
\item Compared to existing methods, the proposed method does not require any offline-training. It can be easily implemented to SLAM system. 
\end{itemize}

This paper is organized as follows: Section \MakeUppercase{\romannumeral 2} reviews the related works on loop closure detection. Section \MakeUppercase{\romannumeral 3} describes the details of the proposed method. Section \MakeUppercase{\romannumeral 4} shows experiment results and comparison with existing works, followed by conclusion in Section \MakeUppercase{\romannumeral 5}.

\section{Related Work}
Most of loop closure detection methods adopt Bag of Words structure which originates from nature language processing. In this model, a text is represented as the multiset of its words, regardless of grammar or word order. Similarly, this idea is applied into loop closure detection such as FABMAP and DBoW2 \cite{cummins2008fab,galvez2012bags}. 
FAB-MAP defines a probabilistic model over the bag-of-words representation \cite{cummins2008fab}. It utilizes Chow-Liu tree \cite{chow1968approximating} to approximate the co-occurrences between SURF feature points. \cite{cummins2009highly} tests datasets of 70 km and 1,000 km in length respectively and achieves a satisfactory recall rate with only a few false positives. DBoW2 \cite{galvez2012bags} creates a tree vocabulary from offline training over a big dataset. New feature points are marked with a sequence number according to the vocabulary so that the co-occurrence of the frames can be estimated by the Euclidean distance of the feature points in the vocabulary. 

Some other research works aim to find an effective and efficient image descriptor for loop closure detection. \cite{mur2017orb} uses SURF feature descriptor for loop detection. It achieves a satisfactory result but consumes lots of computational resource. \cite{huang2016vlad} extracts VLAD vector from each image. VLAD is a first order statistics of the non-probability Fisher Vector \cite{uchida2016image}, which can be obtained by training a codebook of k visual words using k means. The similarity is estimated by measuring the Euclidean distance of related vectors. In recent years, there are also some binary descriptors used in loop closure detection such as Binary Robust Invariant Scalable Keypoints (BRISK) and Binary Robust Independent Elementary Features (BRIEF), \cite{lowe2004distinctive,bay2006surf,viswanathan2009features,calonder2010brief,leutenegger2011brisk}. They take the advantage of fast binary operation and use probability theory to represent features. However, they contain some uncertainty so that the accuracy may drop sometimes.
\begin{figure}[t]
\begin{center}
    \begin{subfigure}{0.49\linewidth}
        \includegraphics[width=0.99\linewidth]{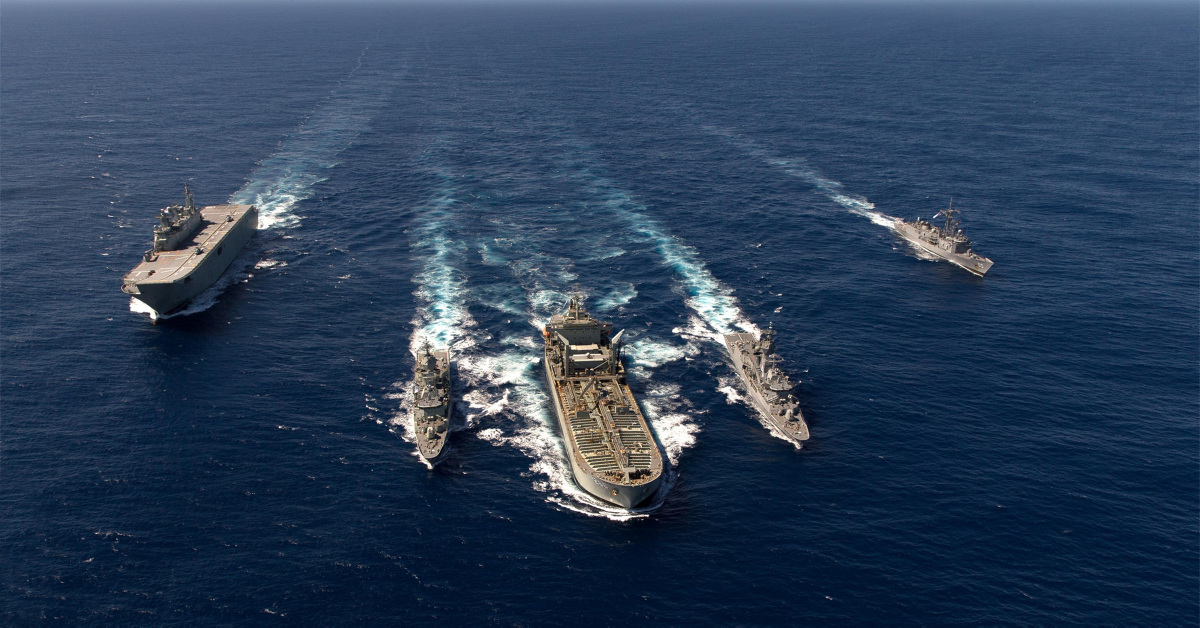}
        \caption{Raw Image}
        \label{fig:Saliency Detection-a}
    \end{subfigure}
    \hfill
    \begin{subfigure}{0.49\linewidth}
        \includegraphics[width=0.99\linewidth]{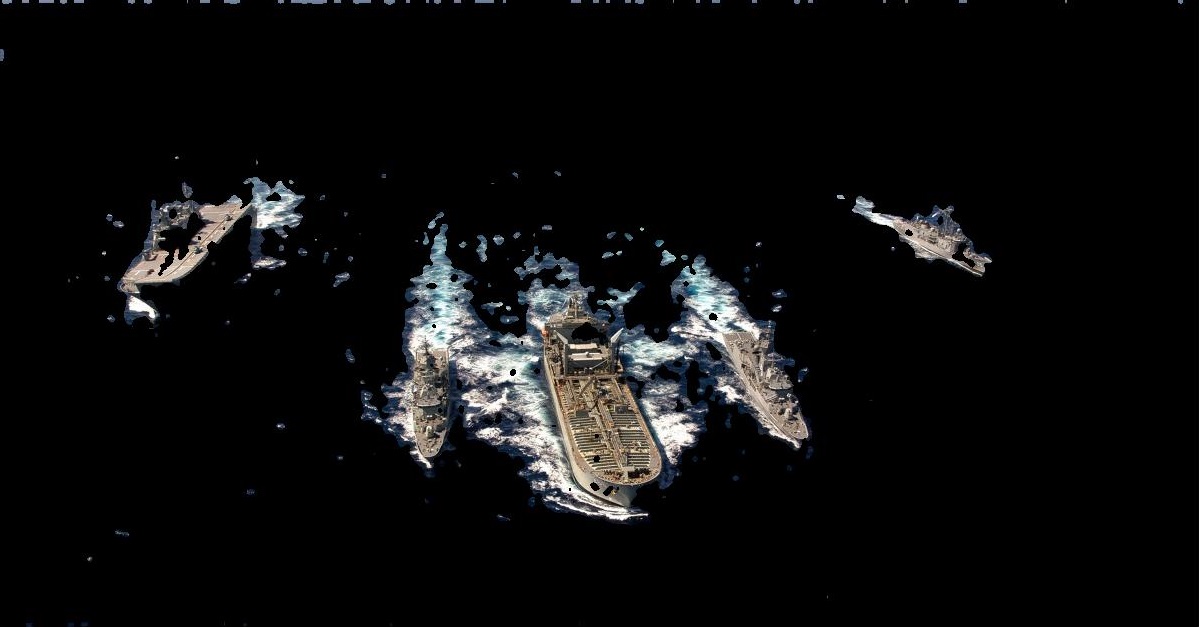}
        \caption{Result}
        \label{fig:Saliency Detection-b}
    \end{subfigure}
\end{center}
\captionsetup{justification=centering}
\caption{Example of Log spectral residual approach.}
\label{fig:Saliency Detection}
\end{figure}

\begin{figure*}[t]
    \begin{subfigure}{0.49\linewidth}
        \begin{center}
        \includegraphics[width=1.0\linewidth]{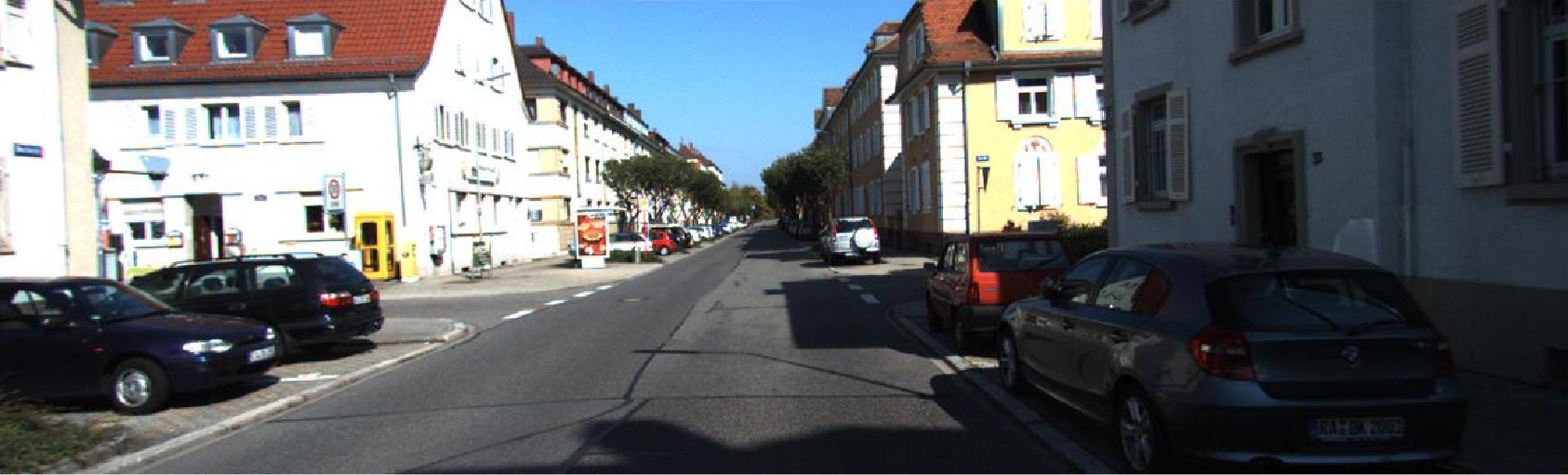}
        \end{center}
        \caption{First frame}
        \label{fig:binaryexample-a}
    \end{subfigure}
    \hfill
    \begin{subfigure}{0.49\linewidth}
        \begin{center}
        \includegraphics[width=1.0\linewidth]{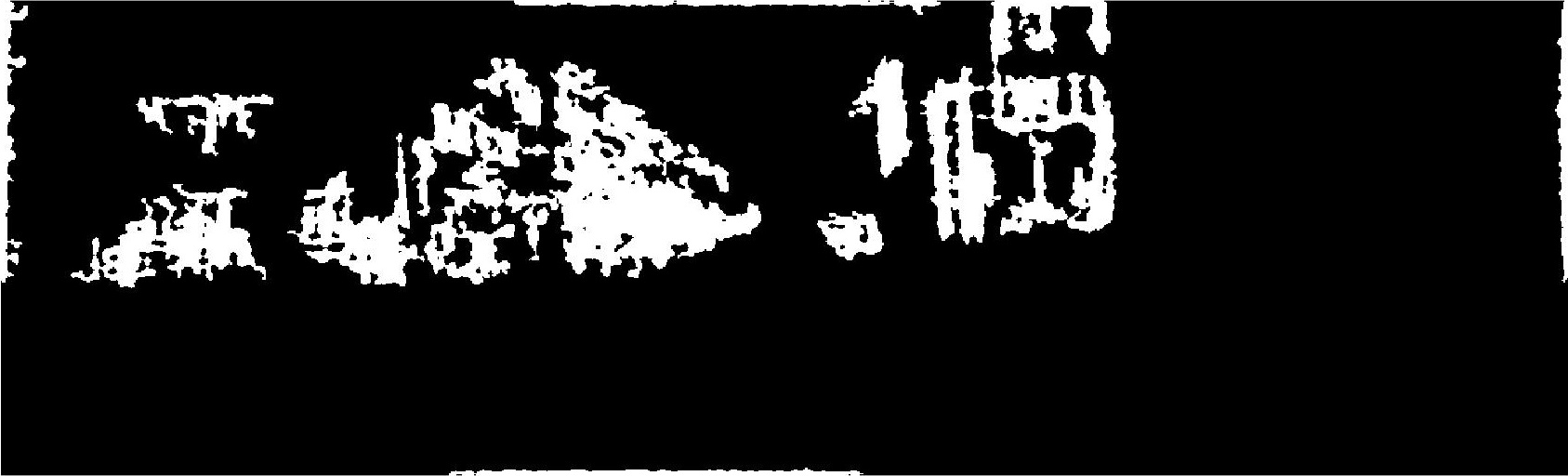}
        \end{center}
        \caption{Binary content extraction result}
        \label{fig:binaryexample-b}
    \end{subfigure}
    \begin{subfigure}{0.49\linewidth}
        \includegraphics[width=0.99\linewidth]{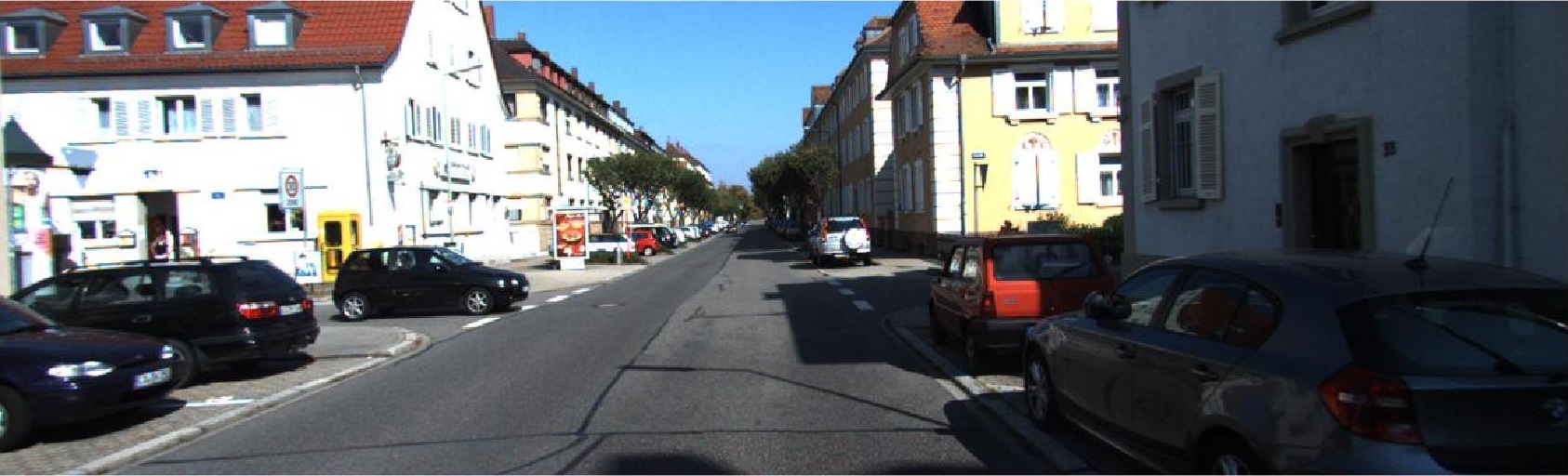}
        \caption{Second frame}
        \label{fig:binaryexample-c}
    \end{subfigure}
    \hfill
    \begin{subfigure}{0.49\linewidth}
        \includegraphics[width=0.99\linewidth]{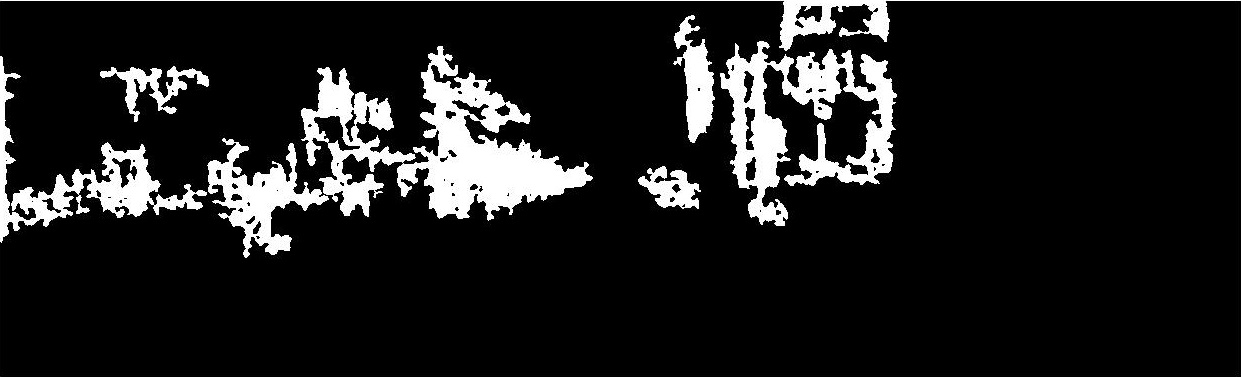}
        \caption{Binary content extraction result}
        \label{fig:binaryexample-d}
    \end{subfigure}
    \begin{subfigure}{0.49\linewidth}
        \includegraphics[width=0.99\linewidth]{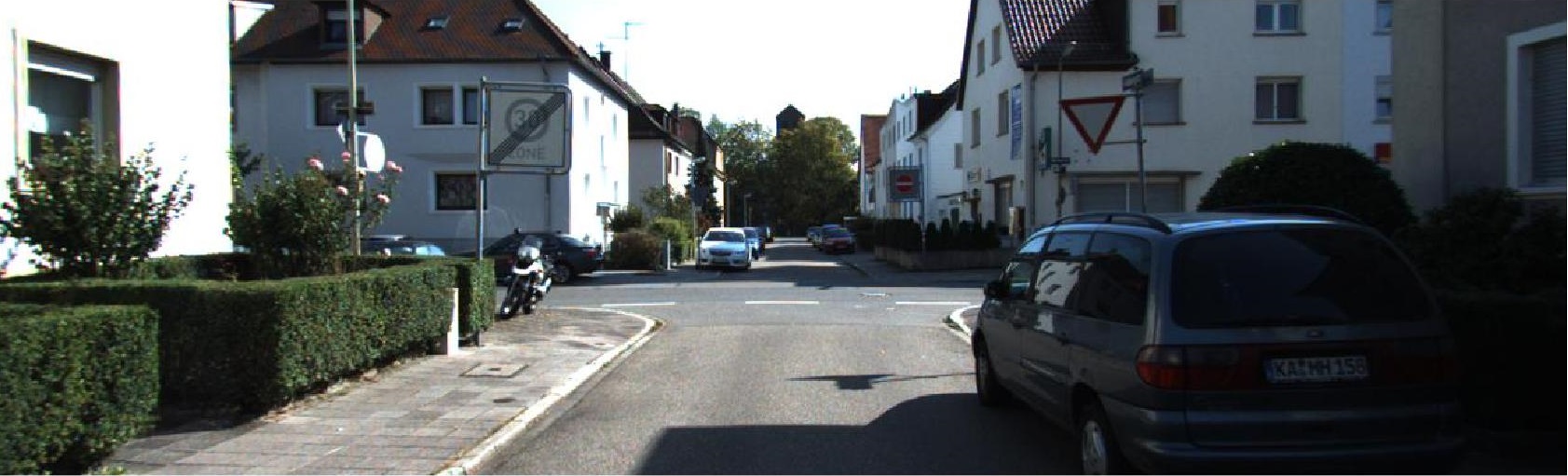}
        \caption{Third frame}
        \label{fig:binaryexample-e}
    \end{subfigure}
    \hfill
    \begin{subfigure}{0.49\linewidth}
        \includegraphics[width=0.99\linewidth]{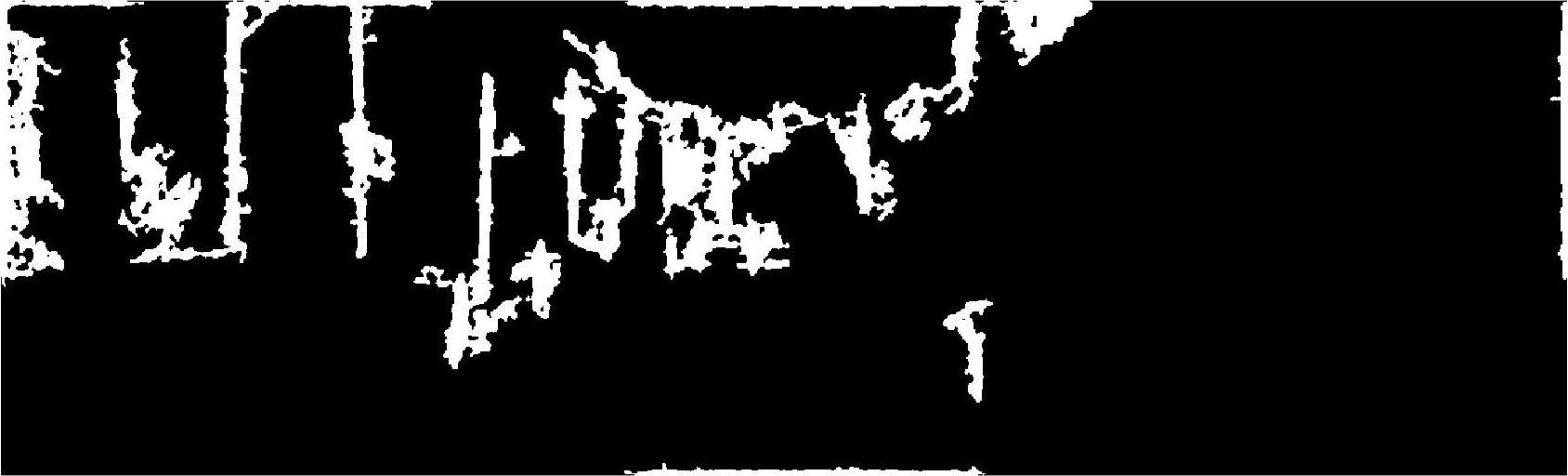}
        \caption{Binary content extraction result}
        \label{fig:binaryexample-f}
    \end{subfigure}
%\captionsetup{justification=centering}
\caption{Examples of binary content extraction.}
\label{fig:binary_extraction_example}
\end{figure*}

Another trend in loop closure detection is the utilization of Deep Learning based descriptors. \cite{chatfield2014return} has conducted a comprehensive evaluation and has shown the advantages of Deep Learning based features. In the work of \cite{hou2015convolutional}, the authors apply a pre-trained Convolutional Neural Network (CNN) model, where the outputs at the intermediate layers are used as image descriptors. The utilization of GPU accelerates the processing speed up to the level of milli-second. In \cite{xia2016loop}, the authors apply PCANet \cite{chan2015pcanet} to extract features as image descriptors. It only takes 10-60 ms on City Center dataset on an NVIDIA GPU with the recall rate up to 20\%. Deep learning method shows a good performance in loop closure detection. However, the application is limited by the requirement of GPU which is costly for robotic systems.

Recently, another research work uses object for loop closure detection \cite{wang2019salient}. It performs loop closure detection based on the objects cropped from each image. It achieves very satisfactory speed but at the sacrifice of recall rate. Another problem is that it can fail if there is any repetitive objects in the scene.

\section{Framework}
%*****************************framework**************************************
The proposed binary content based loop closure detection framework consists of three parts: binary content construction, fast image retrieval and precise loop closure detection, which is shown in Fig. \ref{fig: flowchart}. To utilize the object distribution information, the first step binary content construction extracts the objects or salient regions from the image and then further compresses extracted parts into compact binary image. After that, fast image retrieval performs binary image indexing at high speed and filters out most of unmatched pairs. Lastly, precise loop closure detection conducts further check on the result to remove any false positive. In the process of fast binary content indexing, most of unmatched pairs are filtered out so this process only takes limited computational resources. The details of each step will be explained in this section.

\subsection{Binary Content Construction}
The extracted binary content should be highly representative information of the original image. However, the binary content cannot reveal the color or grey level of pixel so that we only operate at the level of salient region. A salient region generally refers to those image parts that contain rich texture. The location of the salient region and the shape of salient part can be useful for loop detection. Different images will have different salient regions so that it can be a criterion to search for paired images. To extract salient regions, we perform the Log spectral residual method \cite{hou2007saliency}. The Log spectral residual method has the advantage of low computational cost and high extraction capability. Moreover, no prior knowledge is required for this approach.
Generally, given an input image $\mathcal{I}$, we define the following notations:
\begin{itemize}
\item $\mathbf{A}(f)$: The real part of Fast Fourier Transform of image $\mathcal{I}$, $\mathbf{A}(f) = \Re(\mathcal{F}(\mathcal{I}))$.
\item $\mathbf{P}(f)$: The imaginary part of Fast Fourier Transform of image $\mathcal{I}$, $\mathbf{P}(f) = \Im(\mathcal{F}(\mathcal{I}))$.
\item $\mathbf{L}(f)$: The log spectral of $\mathbf{A}(f)$, $\mathbf{L}(f) = \log(\mathcal{A}(f))$.
\end{itemize}
The Log spectral residual $R(f)$ is defined as:
\begin{equation}
\begin{aligned}
R(f) &= L(f) - h_{n}(f) \cdot L(f), \label{Equation:saliecyresult}
\end{aligned}
\end{equation}
where $h_{n}(f)$ is an average filter of an $n\times n$ matrix. Salient region map $\mathcal{O}(x)$ can be derived by recovering equation (\ref{Equation:saliecyresult}) with Gaussian filter $\mathcal{G}(x)$:
\begin{subequations}
    \begin{align}
    S(x) &= \mathcal{G}(x) \cdot \mathcal{F}^{-1}[R(f) + \exp(P(f))]^2,\\
    \mathcal{O}(x) &= 
    \left\{
    \begin{aligned}
    1   &~&~&~\text{if}~ S(x) > E(S(x)) \cdot \gamma, \\
    0   &~&~&~\text{otherwise},
    \end{aligned}
    \right.
    \end{align}
    \label{eqn:salient level}
\end{subequations}
where the threshold $\gamma$ indicates the level of salient region extraction. A larger $\gamma$ implies that less salient area will be ignored, and only highly salient regions or objects will be retained. A demonstration of the Log spectral residual approach is shown in Fig. \ref{fig:Saliency Detection}, where only the crafts are kept after filtering. Salient region contains the most representative information of the image and in most cases it is unique for each image. By binarizing each frame into salient region map and storing it, the database is built up for later processing.

\subsection{Fast Image Retrieval}
Fast image retrieval aims to match binary content with the database. The key idea of this part is to make use of fast logical operation to conduct searching. Similar scenes share similar salient region distribution. When the place is revisited, the light condition or view angle can be slightly changed, but the distribution will remain the same. Hence, by comparing the salient region map $\mathcal{O}_1$ and $\mathcal{O}_2$ we can perform an element-wise similarity check:
\begin{equation}
    \xi = \frac{\mathcal{F}(\mathcal{O}_1 \;\&\; \mathcal{O}_2)}{\max\{\mathcal{F}(\mathcal{O}_1),\mathcal{F}(\mathcal{O}_2\}},
\end{equation}
where $\xi$ is the similarity factor of two images and $\mathcal{F}(\mathbf{x})$ counts the number of "true" values in the matrix. The fast image retrieval can be performed by simply setting threshold to $\xi$. In the meanwhile, we also define a binary image center $\mathcal{M}$
\begin{equation}
    \mathcal{M} = \frac{\sum\limits_{\mathbf{u}\in \mathcal{O}} \mathcal{O}(\mathbf{u}) \cdot \mathbf{u}}{\mathcal{F}(\mathcal{O})},
\end{equation}
where $\mathbf{u}$ is the coordinate of pixel in image. By setting threshold on $\mathcal{M}$ we can simply filter out unmatched pairs. 
An example of binary content based fast indexing is shown in Fig. \ref{fig:binary_extraction_example}. We randomly pick up 3 frames from KITTI dataset \cite{geiger2013vision}. The first and second frames are taken at same place but different time, while the third frame is taken at another similar place. The first and second frames are loop closure pairs but the first and third frames are not. By applying the fast indexing, we can calculate the $\xi$ between frames: $\xi_{12} = 67\%$ and $\xi_{13} = 20\%$. Intuitively we can tell that the second frame is much more similar to the first frame.

\begin{figure}[t]
\begin{center}
    \includegraphics[width=0.99\linewidth]{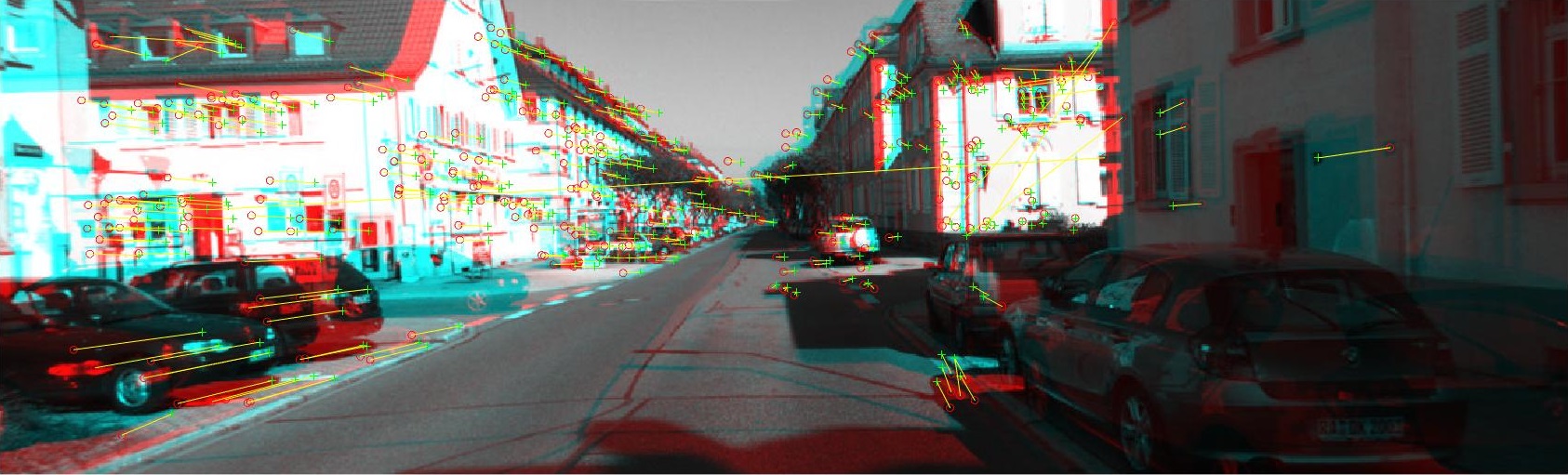}
\end{center}
\caption{An example of SURF feature points matching.}
\label{fig: feature_matching}
\end{figure}
\begin{table}[t]
    \begin{center}
    \begin{tabular}{ccc}
    \toprule
    Dataset & Image Size & Source of Ground Truth \\
    \midrule
    KITTI   &   370$\times$1226  &  GPS   \\ 
    New College     &   640$\times$480        & GPS     \\  
    City Center     &   640$\times$480        &  GPS         \\  
    \bottomrule
    \end{tabular}
    \captionsetup{justification=centering}
    \caption{Information of Different Datasets.}
    \label{table:datasets information.}
    \end{center}
\end{table}

\begin{table}[t]
\begin{center}
\begin{tabular}{cccc}
\toprule
Dataset  & Mean Time & Average Recall Rate & Precision \\ 
\midrule
KITTI    &  130        &         54.9            & 100       \\ 
New College     &  92       &         20.9           & 100      \\  
City Center      &  86        &         27.7           & 100      \\  
\bottomrule
\end{tabular}
%\captionsetup{justification=centering}
\caption{Loop detection results of our approach.}
\label{table:Experiment results on different dataset.}
\end{center}
\end{table}

\begin{figure}[h]
\begin{center}
    \begin{subfigure}{0.70\linewidth}
        \includegraphics[width=1.00\linewidth]{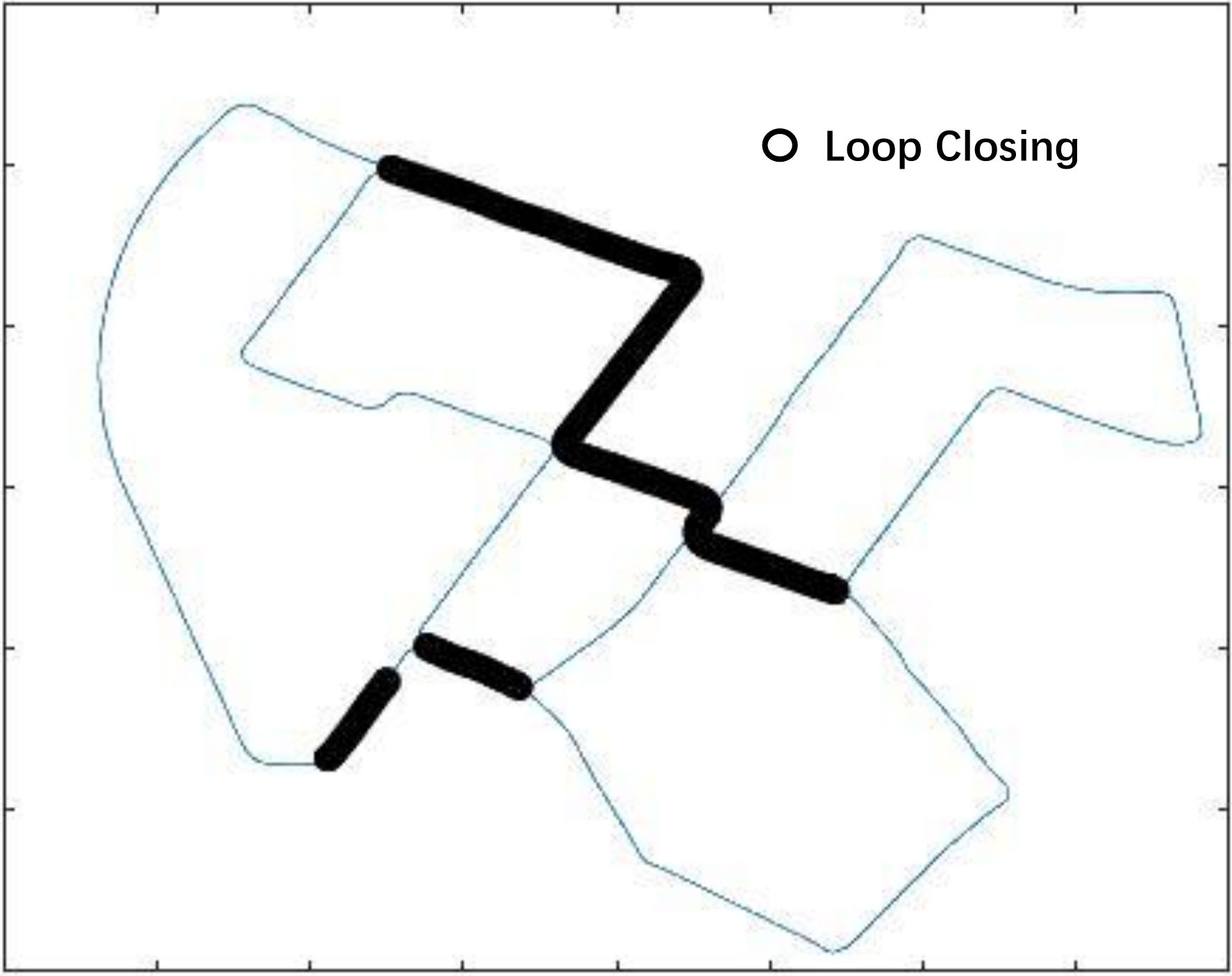}
        \captionsetup{justification=centering}
        \caption{Ground truth of KITTI sequence 00.}
        \label{fig:kitti_result-a}
    \end{subfigure}
    \hfill
    \begin{subfigure}{0.70\linewidth}
        \includegraphics[width=1.00\linewidth]{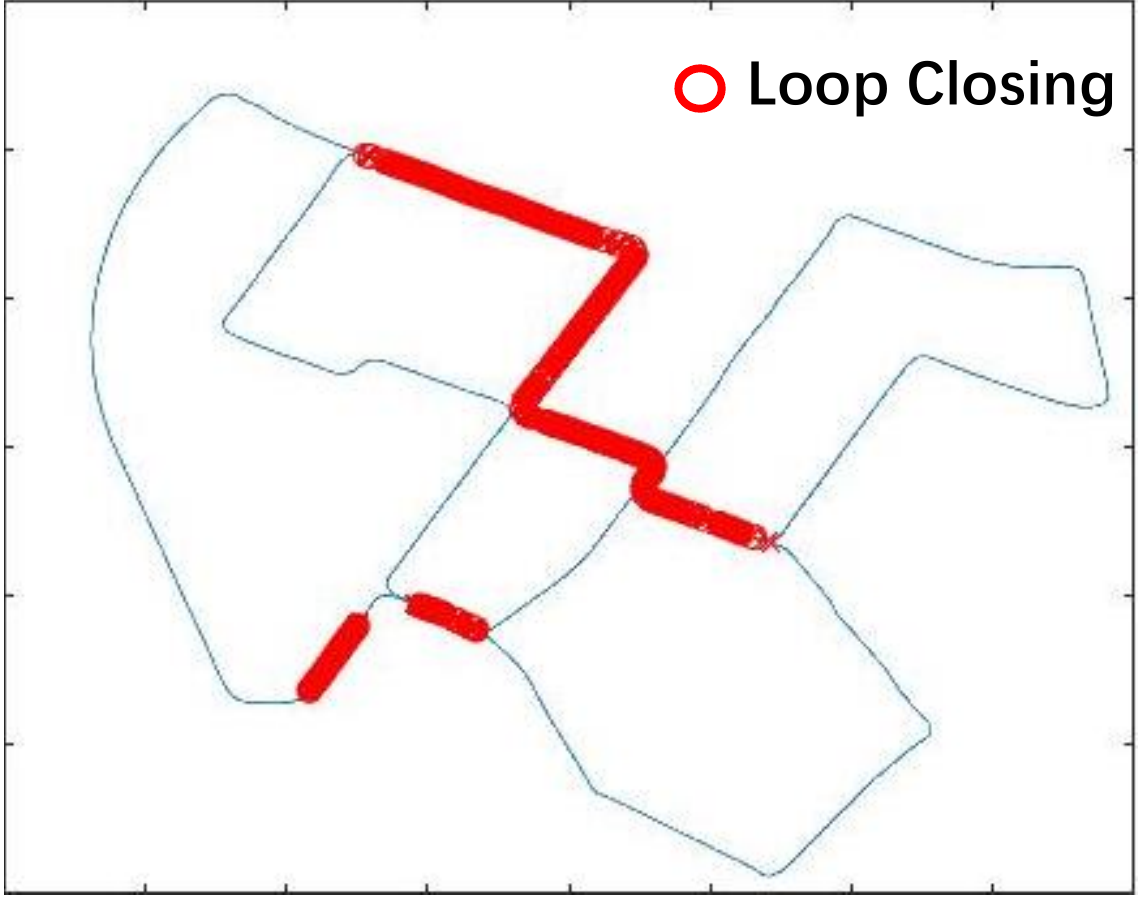}
        \captionsetup{justification=centering}
        \caption{Loop closure detection result.}
        \label{fig:kitti_result-b}
    \end{subfigure}
\end{center}
\captionsetup{justification=centering}
\caption{Loop closure detection result of the proposed method.}
\label{fig:kitti_result}
\end{figure}
\subsection{Precise Loop Closure Detection}
The fast image retrieval is able to remove most unmatched pairs. However, the binary content only considers the structure of the content which is fast but not accurate enough. Considering that traditional method using SURF feature descriptor has a good performance in matching images, we can implement feature point based comparison to further increase the precision.

The SURF feature points are extracted from each frame due to its high precision in image matching \cite{bay2006surf}. And we use SURF descriptors to examine each image pair. Fig. \ref{fig: feature_matching} shows an example of feature matching. The number of matched feature points reveals the similarity of image pair.

\section{Experiment Results}

To prove its robustness, we test the proposed method with different datasets including KITTI dataset, New College dataset and City Center dataset \cite{geiger2013vision,smith2009new,engel2016photometrically}. The information of respective dataset is given in Table \ref{table:datasets information.}. The most important performance indexes for loop closure detection are recall rate, recall precision and speed. Recall precision refers to the ratio of correct loop closure detection against total loop closure detected. The higher recall precision the better, since any false positive may cause filter divergence easily. Recall rate refers to the number of correct loop pairs detected against total loop pairs which can be collected from the ground truth. In this section, we provide a detailed analysis of our proposed method.

\begin{figure*}[t]
    \begin{subfigure}{0.49\linewidth}
        \begin{center}
        \includegraphics[width=0.8\linewidth]{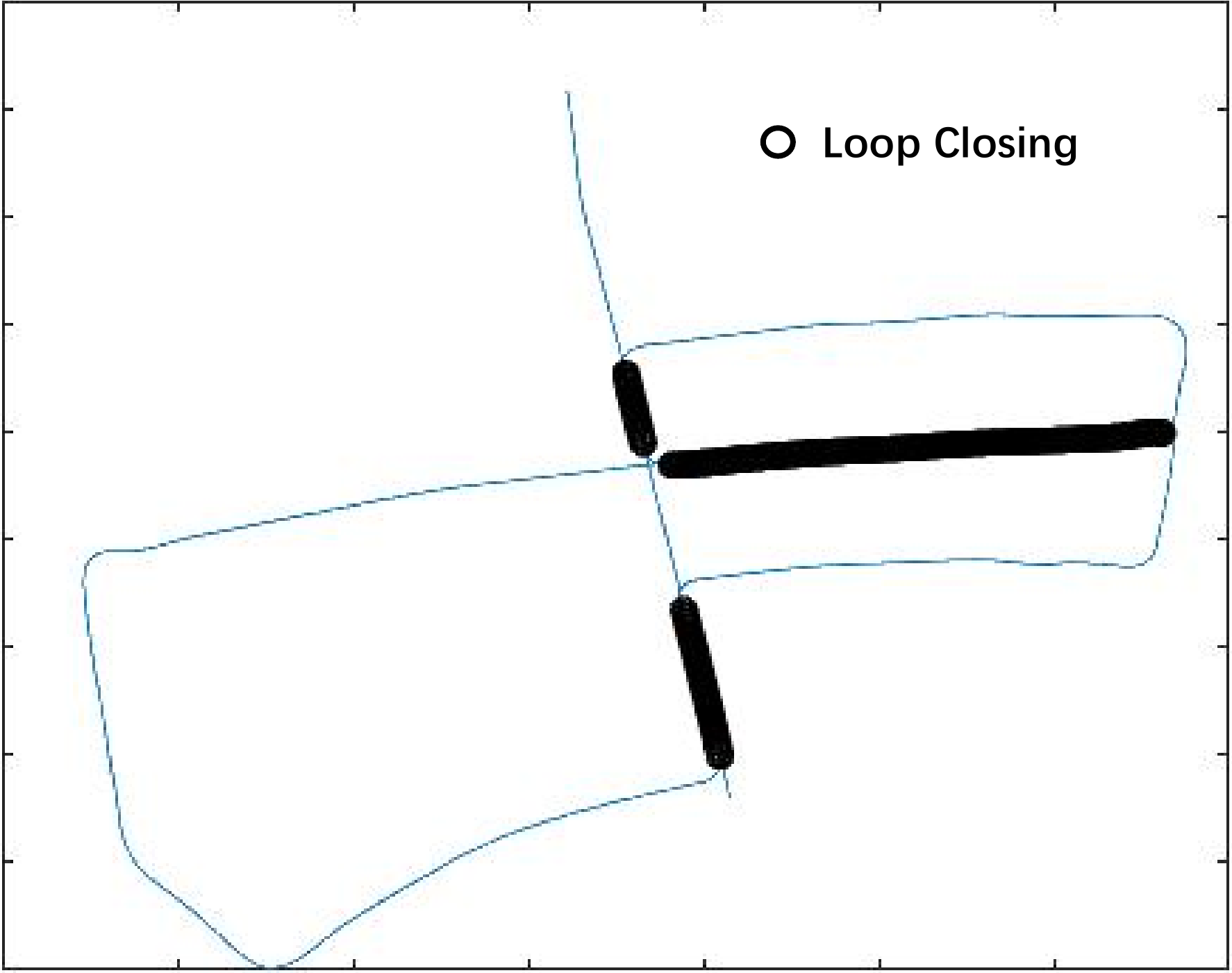}
        \end{center}
        \caption{Ground truth}
        \label{fig:result_comparison-a}
    \end{subfigure}
    \hfill
    \begin{subfigure}{0.49\linewidth}
        \begin{center}
        \includegraphics[width=0.8\linewidth]{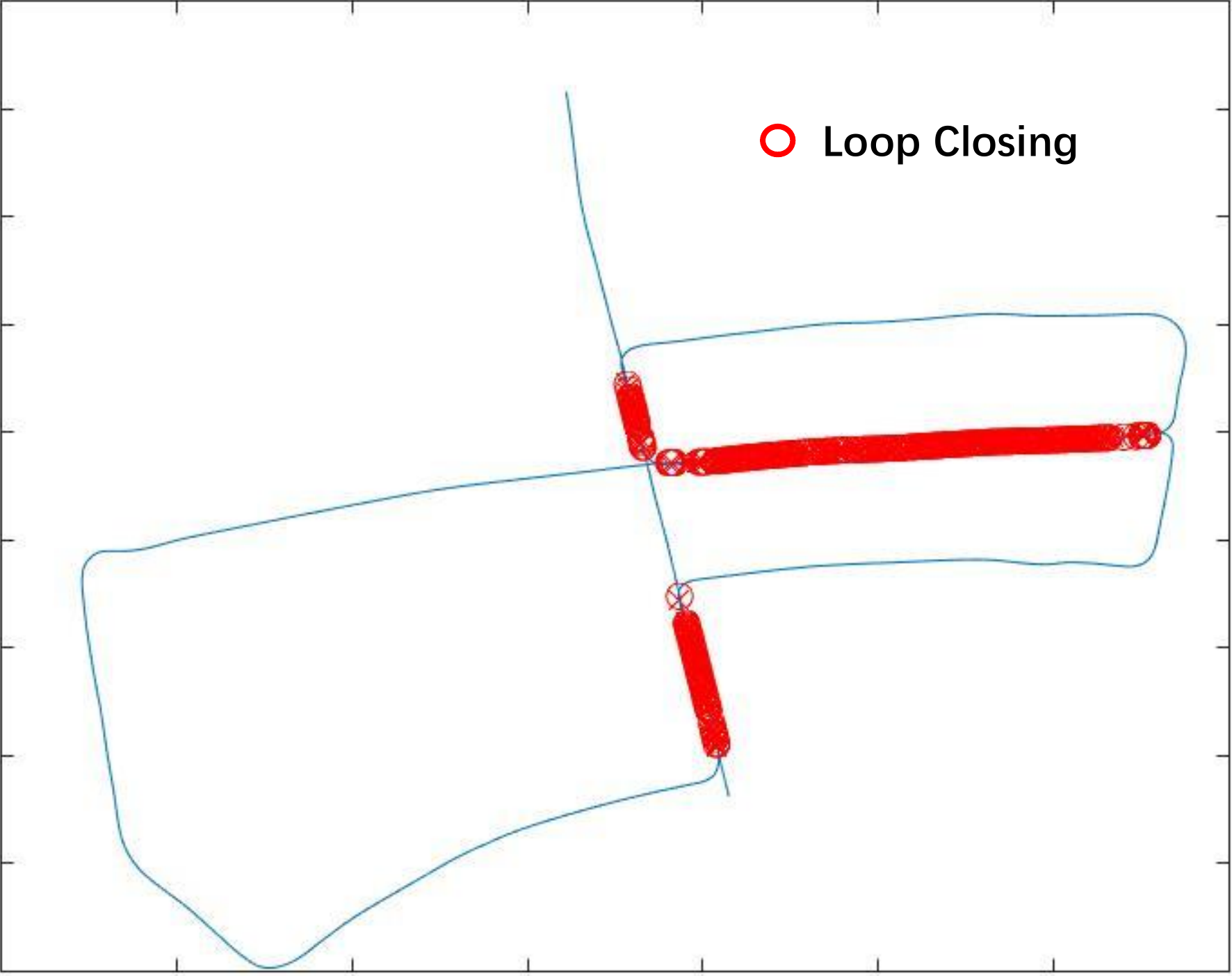}
        \end{center}
        \caption{Binary content-based approach}
        \label{fig:result_comparison-b}
    \end{subfigure}
    \hfill
    \begin{subfigure}{0.49\linewidth}
        \begin{center}
        \includegraphics[width=0.8\linewidth]{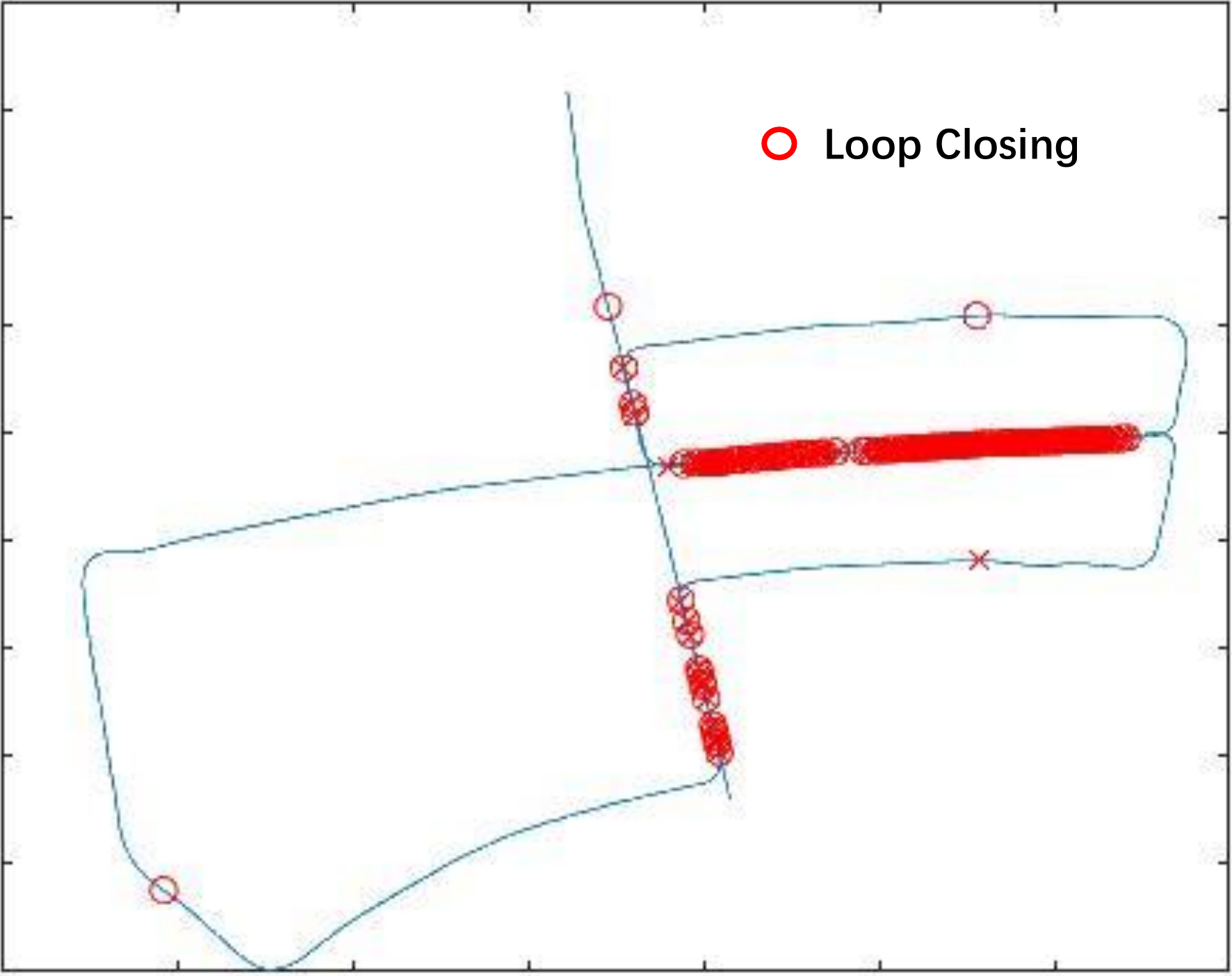}
        \end{center}
        \caption{FABMAP}
        \label{fig:result_comparison-c}
    \end{subfigure}
    \hfill
    \begin{subfigure}{0.49\linewidth}
        \begin{center}
        \includegraphics[width=0.8\linewidth]{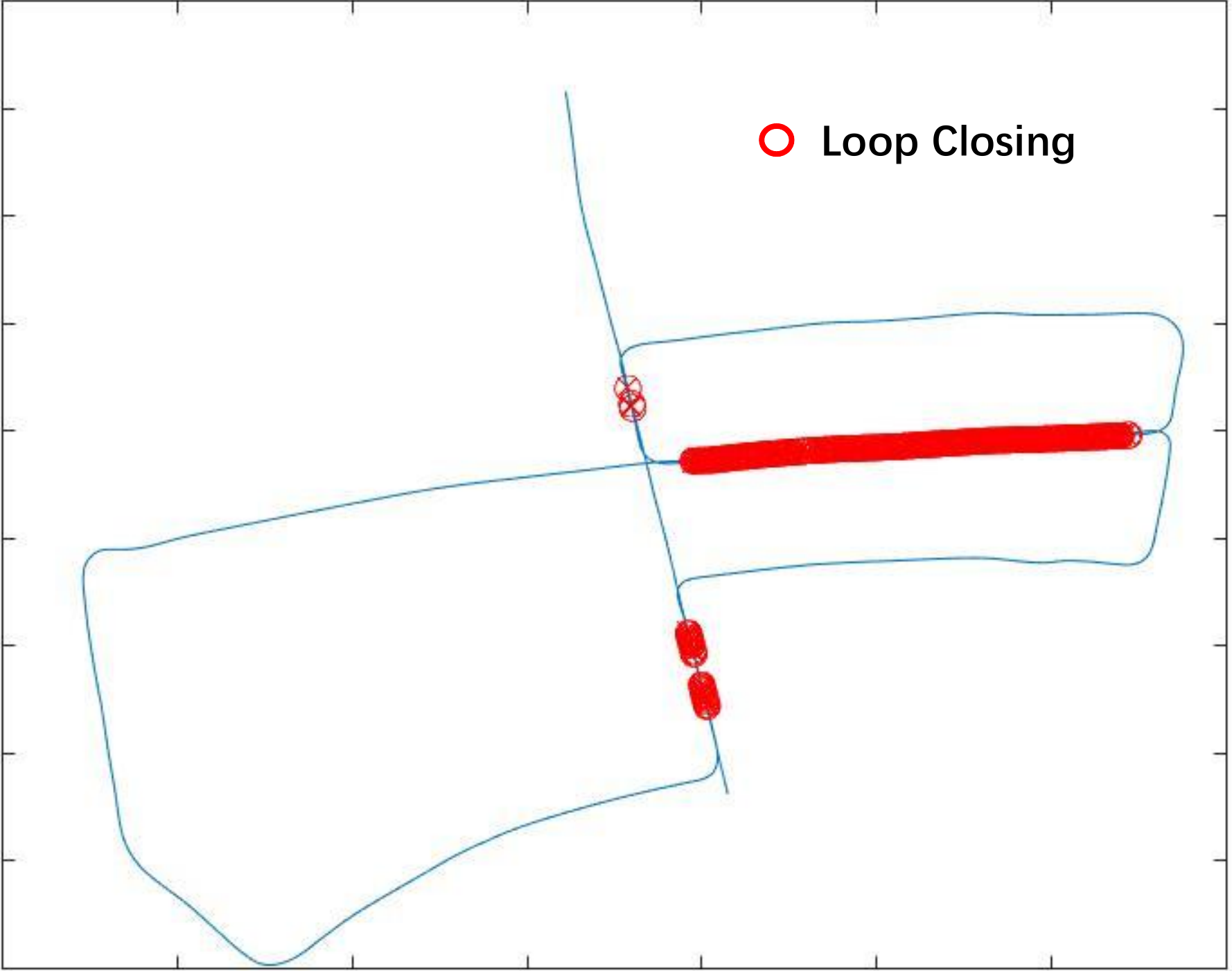}
        \caption{DBoW2}
        \label{fig:result_comparison-d}
        \end{center}
    \end{subfigure}

%\captionsetup{justification=centering}
\caption{Comparison of binary content extraction with existing methods.}
\label{fig:result_comparison}
\end{figure*}

\begin{table*}[t]
\setlength{\tabcolsep}{2pt}
\begin{center}
\begin{tabular}{c|ccc|ccc|ccc}
\toprule
\multirow{2}{*}{Dataset}         & \multicolumn{3}{c|}{Sequence 00}   & \multicolumn{3}{c|}{Sequence 02}       &\multicolumn{3}{c}{Sequence 05}          \\ 
         & Mean Time (ms) & Recall Rate (\%) & Precision (\%) & Mean Time & Recall Rate & Precision  & Mean Time & Recall Rate & Precision  \\ 
\midrule
\multicolumn{1}{c|}{Our Approach}   & 130        & 54.9   & 100                                                            & 129        & 47.7     & 100  
                                    & 118        & 62.5     & 100  \\ 
\multicolumn{1}{c|}{FABMAP} & 1124         & 32.2            &   97.7        
                            & 1162           &     23.4         & 49
                            & 1021           &     35.3         & 98\\ 
\multicolumn{1}{c|}{DBoW2}       & 460         & 57.2            &   100     
                                & 448            &     38.9          & 100  
                                & 355            &     54.0          & 100  \\ 
\bottomrule
\end{tabular}
\caption{Quality Analysis of FABMAP and the proposed method on different datasets.}
\label{table:comparison of different methods}
\end{center}
\end{table*}
\subsection{Experiment result on public dataset}
We conduct the test on an intel\textregistered\;NUC mini computer which is popularly used in robotics relateed applications. The proposed method is tested with different datasets mentioned above. The loop closure detection results are collected and displayed in Matlab for visualization purpose. An example of our loop closure detection approach on KITTI dataset is shown in Fig. \ref{fig:kitti_result}. In the figure we plot the moving trajectory of camera and mark the ground truth of loop closure detection with black circle on the first image, while the detection result is shown on the second image with red circle marked instead. Each circle refers to a loop closure pair. Intuitively we can tell that there is no false positive detected and most of loop closure places are identified. Our proposed method achieves a recall rate of 54.9\% and a recall precision of 100\% which is very satisfactory. More test results can be found in Table \ref{table:Experiment results on different dataset.}. In total we pick up 5 recordings with loop closure from KITTI dataset, and our method achieves more than 50\% on average without any false positive. It also achieves 20\% on New College dataset and 27\% on City Center dataset without false positive. In the meanwhile, our methods still can run at high speed of 10 Hz on average. 

\subsection{Comparison with other methods}

We further compare our method with the state-of-the-art methods such as FABMAP, DBoW2 \cite{cummins2008fab,galvez2012bags}. To be consistent, all experiments are conducted on an intel\textregistered\;NUC mini computer. In order to have a clear comparison, we pick the largest datasets with loop closure for demonstration since the efficiency differs more as database size increases. We use KITTI sequence 00, KITTI sequence 02 and KITTI sequence 05 with more than 10k frames in total. We test KITTI sequence 05 on each method first and the result is shown in Fig. \ref{fig:result_comparison}. In the experiment, we finely tuned the threshold in both FABMAP and DBoW2 in order to get the best recall rate and recall precision. However, our approach does not require to tune any parameter for specific dataset. Besides, both FABMAP and DBoW2 require offline training of similar dataset in advance, while the proposed method does not. Our method reports most of the loop closure places correctly while FABMAP has false positive and DBoW2 fails to report loop closure in some places. The details of the rest results on other datasets are shown in Table \ref{table:comparison of different methods}. The speed of the proposed method is 3 times faster than DBoW2 and 9 times faster than FABMAP. In our approach, we use the sophisticated SURF feature to achieve the precision because feature-wise comparison does not occur frequently. Hence our approach also provides reliable precision and recall rate. A demonstration of the experiment result can be found at \url{https://youtu.be/YCRd3N0LwSA}.

\section{Conclusion}
%conclusion 的部分，可以讲一下有关于复用性的问题，比如第一部分fft可以在之后用到，三个指标也可以在之后用到
In this paper, we have presented a fast loop closure detection method via binary content. Traditional approaches such as FABMAP and DBoW2 use feature descriptors to compress the image content and build a descriptor vocabulary for indexing. However, these methods require intensive mathematical calculation to estimate the similarity of two images, which is less efficient than binary operation. Observe that operation on binary image can have a similar result but at higher speed than feature descriptor. Hence based on the observation, we proposed a new framework for loop closure detection which consists of three parts: binary content construction, fast image retrieval and precise loop closure detection. The experiment result has demonstrated that it is able to detect most of loop closure places without false positive. The proposed method was also compared with state-of-the-art methods such as FAB and DBoW2. The result has shown that it outperforms other approaches in both recall rate and speed. In addition, no offline training is required in our approach so that it is easy for implementation.

\section*{ACKNOWLEDGMENT}
The author would like to thank Mr. Wang Chen for many great suggestions during the course of this research work.
\balance
%\newpage
\bibliographystyle{IEEEtran}
\bibliography{IEEEabrv,references}

\end{document}